\pdfoutput=1

\documentclass[11pt]{article}

\usepackage[]{ACL2023}

\usepackage{times}
\usepackage{latexsym}
\usepackage{xspace}
\usepackage{booktabs}       
\usepackage{pgfplots}
\usepackage{multirow}
\usepackage{amssymb}
\usepackage{tabularx}
\usepackage{color}
\usepackage{colortbl}

\usepackage[T1]{fontenc}

\usepackage[utf8]{inputenc}

\usepackage{microtype}

\usepackage{inconsolata}

\newcommand{\dataset}{\textsc{DeFacto}\xspace}

\title{On Improving Summarization Factual Consistency  from \\ Natural Language Feedback}

\author{Yixin Liu\Thanks{~Most of the work was done while the first author was an
intern at Microsoft Research.} $^{1}$, Budhaditya Deb$^{2}$, Milagro Teruel$^{2}$,
\\ \textbf{Aaron Halfaker}$^{2}$\textbf{,} \textbf{Dragomir Radev}$^{1}$\textbf{,} \textbf{Ahmed H. Awadallah}$^{2}$\\
$^1$Yale University, $^2$Microsoft Research \\
\texttt{\{yixin.liu, dragomir.radev\}@yale.edu},
\texttt{\{Budha.Deb, hassanam\}@microsoft.com}
}

\begin{document}
\maketitle

\begin{abstract}
 Despite the recent progress in language generation models, their outputs may not always meet user expectations.
 In this work, we study whether informational feedback in natural language can be leveraged to improve generation quality and user preference alignment.
 To this end, we consider \textit{factual consistency} in summarization, the quality that the summary should only contain information supported by the input documents, as the user-expected preference.
 We collect a high-quality dataset, \textbf{DeFacto}, containing human demonstrations and informational natural language feedback consisting of corrective instructions, edited summaries, and explanations with respect to the factual consistency of the summary. 
 Using our dataset, we study three natural language generation tasks: (1) \textit{editing a summary} by following the human feedback, (2)\textit{ generating human feedback} for editing the original summary, and (3) \textit{revising the initial summary} to correct factual errors by generating both the human feedback and edited summary. 
 We show that DeFacto can provide factually consistent human-edited summaries and further insights into summarization factual consistency thanks to its informational natural language feedback. 
 We further demonstrate that fine-tuned language models can leverage our dataset to improve the summary factual consistency, while large language models lack the zero-shot learning ability in our proposed tasks that require controllable text generation. 
\end{abstract}

\section{Introduction}

\begin{figure}[t!]
    \centering
    \includegraphics[width=1.0\linewidth]{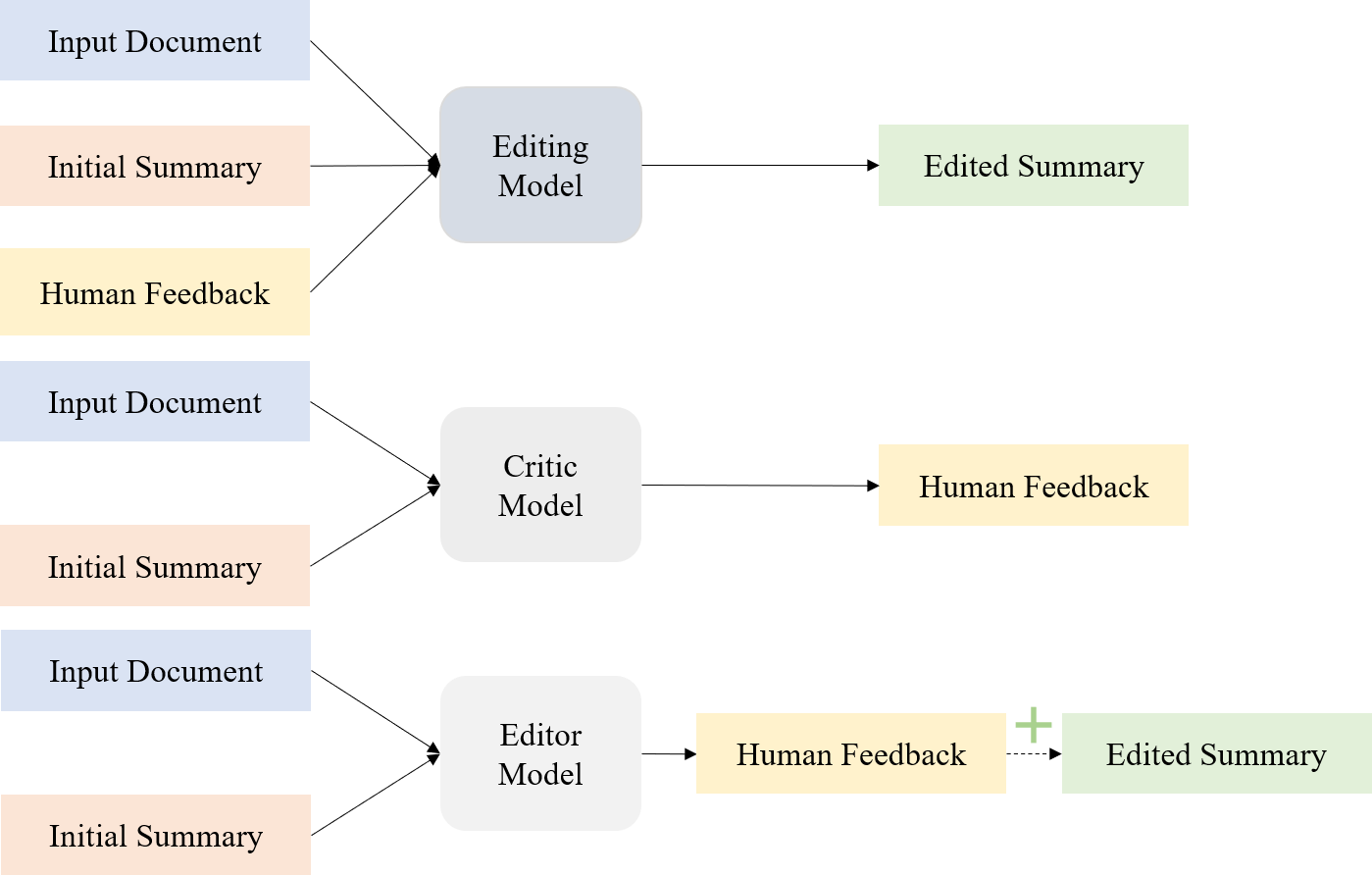}
    \caption{Three NLG tasks studied using our dataset. 
    The \textbf{\textit{Editing}} model aims to improve the initial system-generated summary given human feedback.
    The \textbf{\textit{Critic}} model aims to predict human feedback according to a user-required quality.
    The \textbf{\textit{Editor}} model aims to automatically correct factual errors by predicting both the human feedback and edited summary. 
    }
    \label{fig:intro}
\end{figure}

While recent natural language generation (NLG) models~\cite{radford2019language, lewis-etal-2020-bart, JMLR:v21:20-074, NEURIPS2020_1457c0d6} have made significant progress on the generation quality, they cannot always generate outputs that meet the user needs.
For example, while state-of-the-art summarization systems can generate fluent and relevant summaries, recent work~\citep{goyal-durrett-2021-annotating, Tang2022UnderstandingFE} have shown that they still make errors on fine-grained qualities such as \textit{factual consistency}.\footnote{Following \citet{goyal-durrett-2021-annotating}, we define factual consistency as the summary quality that \textit{all the information of the summary can be supported by the source document}.}
These errors can lead to serious risks to the intended users and make it difficult for them to trust the systems for their decision-making.

\begin{figure*}[t!]
    \centering
    \includegraphics[width=0.95\linewidth]{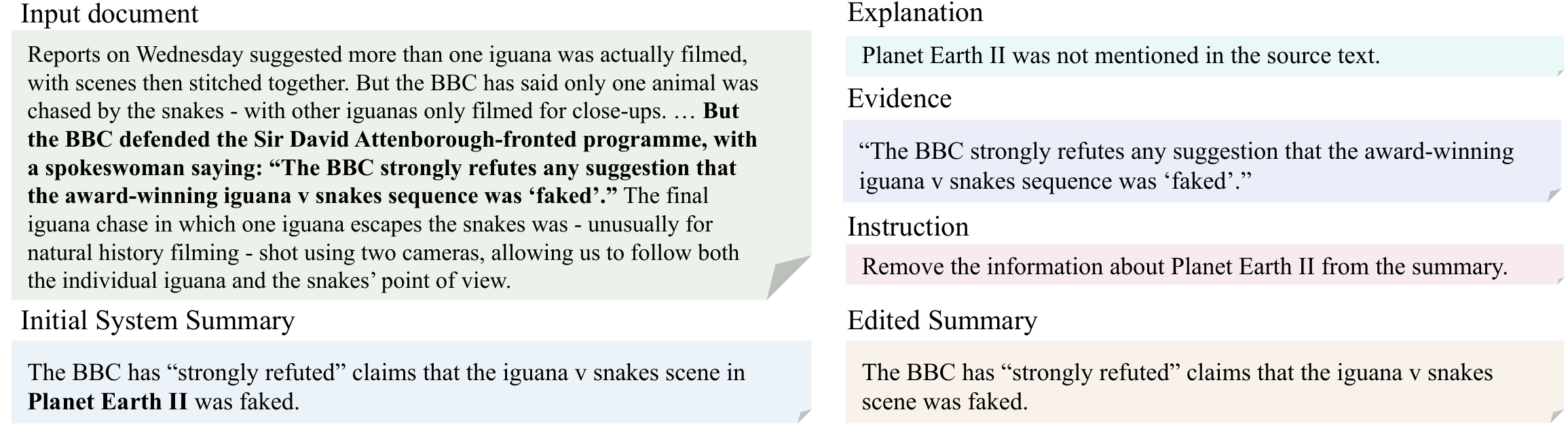}
    \caption{Data example in \dataset. The initial summary contains a factual error about the name of the program \textit{Planet Earth II}. The annotator provided an \textit{\textbf{explanation}} about why the initial summary is not factually consistent, \textit{\textbf{evidence}} (i.e., a sentence in the input document) to support their claims, \textit{\textbf{instructions}} on how to correct the summary, and an \textit{\textbf{edited summary}} (a demonstration) without the factual error.}
    \label{fig:example}
\end{figure*}

Such failures in satisfying the user needs have an \textit{intrinsic} reason -- the large benchmark datasets that are used to train NLG models are usually not collected according to pre-defined user needs, which results in a discrepancy between \textbf{model behaviors} and \textbf{user expectations}.
For example, XSum~\citep{narayan-etal-2018-dont}, one of the most commonly used summarization datasets, contains a large portion of reference summaries with \textit{hallucinations}.\footnote{\citet{maynez-etal-2020-faithfulness} reports that around 76.9\% reference summaries on the XSum dataset contains hallucinated contents that are \textit{not} supported by the source documents.} 
As a result, summarization models trained on XSum dataset generate many non-factual contents, more than models trained on datasets such as CNN/DailyMail~\cite{10.5555/2969239.2969428} dataset~\cite{goyal-durrett-2021-annotating}.
Unfortunately, it can be prohibitively expensive to collect new, large-enough datasets to train NLG models according to user needs, as they can be diverse, personal, and ever-changing over time.

Instead of aligning an existing NLG model to a specific user need, we explore adjusting \textit{model outputs} according to the user needs through \textbf{human demonstrations and feedback}. 
Specifically, we investigate three scenarios (Fig.~\ref{fig:intro}): 
(1) an \textbf{\textit{Editing}} model that aligns initial system outputs to human demonstrations based on the user feedback; 
(2) a \textbf{\textit{Critic}} model that predicts user feedback of initial system outputs according to the user requirements;
(3) an \textbf{\textit{Editor}} model that automatically aligns the initial system outputs to user needs by predicting both the user feedback and edited summary.

We choose \textbf{\textit{factual consistency}} of system-generated summaries as the \textit{user-required quality} to study the aforementioned application scenarios.
To this end, we collect a high-quality, informational dataset containing human demonstrations and feedback.
Specifically, the annotators are presented with initial system-generated summaries and asked to make changes to the summaries to make them factually consistent if they find errors in them. 
Apart from the \textbf{human-edited, factually consistent summaries}, the annotators are also required to provide \textbf{instructions} on how to change the initial summaries (i.e., if they find errors in them) and \textbf{explanation} on why the initial summaries are factually consistent or not.
An example of our dataset is shown in Fig.~\ref{fig:example}.
Using the collected dataset, we show that (1) the \textit{Editing} model can effectively leverage human feedback to adjust the initial system outputs towards human demonstrations;
(2) the \textit{Critic} model can learn to generate meaningful feedback that can be used by the \textit{Editing} model;
(3) the \textit{Editor} model can automatically correct factuality errors without explicit human intervention.
Moreover, we find that the \textit{Editor} model achieves better performance than the baseline model that only generates the edited summary, which indicates that natural language feedback can be beneficial for training models for the corresponding task.

Our contributions can be briefly summarized as: 
(1) we collect \textbf{DeFacto},\footnote{We make the \textbf{DeFacto} dataset publicly available at \url{https://github.com/microsoft/DeFacto}.} a \textit{high-quality dataset} containing human \textbf{De}monstrations and \textbf{F}eedback for improving f\textbf{act}ual c\textbf{o}nsistency of text summarization;
(2) we conduct comprehensive analyses on the collected dataset, which provides further insights about factual consistency in text summarization, such as the relation between the type of factual errors and the type of editing operations;
(3) we provide strong baseline models for the proposed three NLG tasks -- summary editing (\textit{Editing} model), feedback generation (\textit{Critic} model), and automatic factuality error correction with feedback prediction (\textit{Editor} model), which illustrates methods of leveraging natural language feedback for aligning model outputs with user expectations. 
(4) we present two case studies with large language models (LLMs) such as GPT-3.5~\citep{ouyang2022training}, showing that LLMs still lack the \textit{controllable} text generation ability in our proposed tasks.

\section{The \dataset Dataset}

Our dataset, \dataset, contains human demonstrations and feedback w.r.t. the factual consistency of system-generated summaries.
We choose \textbf{XSum dataset} as the target dataset to conduct the data collection because it is the most commonly studied dataset for summarization factual consistency. 
For the system-generated summaries, we select \textbf{PEGASUS}~\citep{10.5555/3524938.3525989}, a top-performing summarization model to generate summaries on both the validation and test set of the XSum dataset.

\subsection{Annotation Process}
\label{subsec:annotation}
Our annotation process follows the following steps:

\noindent (1) \textbf{Detect errors}: The annotator is required to evaluate a summary given the source document and \textbf{decide if the summary is factually consistent}.

\noindent (2) \textbf{Categorize errors}: If the annotator decides the summary \textit{is not} factually consistent, they are required to \textbf{categorize the factual errors} in the summary as either \textit{intrinsic} or \textit{extrinsic}.\footnote{Following \citet{goyal-durrett-2021-annotating}, we define \textbf{intrinsic errors} as errors that \textit{arise as a result of misinterpreting information from the source article} and \textbf{extrinsic errors} as errors that \textit{hallucinate new information or facts not present in the source article}.}
We note that both error detection and categorization are defined at the summary level.

\noindent (3) \textbf{Give explanation}: The annotator is required to \textbf{provide a natural language explanation} on why the summary is factually consistent or not.

\noindent (4) \textbf{Provide evidence}: The annotator is required to \textbf{select a sentence from the source document as evidence} to support their claims described in (3).

\noindent (5) \textbf{Write corrective instruction}: The annotator is required to \textbf{provide instructions} of how to correct the original summary if they think it is not factually consistent.
To enforce uniformity and reduce the noise in the instructions, we provide six templates for the annotators corresponding to different operations: \textit{Remove}, \textit{Add}, \textit{Replace}, \textit{Modify}, \textit{Rewrite}, and \textit{Others}.
The annotators need to fill in the templates to generate the instructions.
The details of the templates are in Appendix~\ref{subsec:ins-temp}. 

\noindent (6) \textbf{Correct summary}: Following the instruction in (5), the annotator is required to \textbf{edit the initial summary} to make it \textit{factually consistent} with minimal, necessary modifications.

\noindent We provide annotated examples in Appendix~\ref{subsec:examples}.

\subsection{Data Collection}

We conduct our data collection on Amazon Mechanical Turk\footnote{\url{https://www.mturk.com/}} (MTurk) platform.
The MTurk annotators need to pass a qualification test to be able to accept our assignments. 
The qualification test includes three actual annotation tasks, and we manually checked the correctness of the answers of the annotators and assigned them scores accordingly. 

For the actual tasks, we collected one annotation for each example (i.e., a document-summary pair), and collected around 1000 examples on the test set and 1500 examples on the validation set.
To estimate the inter-annotator agreement, we additionally collect two more annotations for 100 examples on the test set.
We require the annotators to be located in the United States.
Depending on the difficulty of the assignments, %
the annotators are compensated with 1.2 - 2.0 US dollars per assignment accordingly based on a \$12/hour pay rate. 

To check the inter-annotator agreement on steps (1) \textit{Detect Errors} and (2) \textit{Categorize Errors} in \S\ref{subsec:annotation}, we calculated the Krippendorff's alpha~\citep{Krippendorff2011ComputingKA}, and found that the agreement score is 0.5552, 0.1899, 0.5260 for if the summary contains \textbf{\textit{extrinsic}} factual errors, \textbf{\textit{intrinsic}} factual errors and \textbf{\textit{any}} factual errors respectively.
For human-written \textbf{\textit{explanation}}, \textbf{\textit{instructions}}, and \textbf{\textit{edited summary}} in step (3), (5), (6) in \S\ref{subsec:annotation}, we calculated the ROUGE~\citep{lin-2004-rouge} score among the answers provided by different annotators, and found the average ROUGE-1 F-score to be 30.52, 50.96, 71.77, respectively. 
Lastly, for the evidence in step (4), we consider two sentences as equivalent if the ROUGE-1 score between them is above 90.  
We found the match rate among different annotators to be 0.4403.

\section{\dataset Analysis}

\begin{table}[t!]
\centering
\small
\begin{tabular}{lcccc}
\toprule
 & \textbf{Train} & \textbf{Val} & \textbf{Test} & \textbf{All} \\
\midrule
\textbf{All} &  1000 &  486  &  1075 & 2561 \\
\textbf{w/ Errors} & 701  & 341  &  779 & 1821 \\
\bottomrule
\end{tabular}
\caption{Numbers of data points in \dataset dataset.
71.1\% of annotated summaries contain factual errors.
}
\label{tab:basic} 
\end{table}

With the collected annotations, we further split the data collected on the validation set of XSum dataset into a training set and a validation set for the following experiments.
We perform data analyses with different aspects of the collected dataset.
The basic dataset statistics are in Tab. \ref{tab:basic}.
Out of all the examples, \textbf{71.1\%} of them contain at least one factual error, \textbf{58.8\%} of them contain \textit{extrinsic errors}, \textbf{22.0\%} of them contain \textit{intrinsic errors}, and \textbf{9.63\%} of them contain \textit{both} types of errors.

\subsection{Edited Summary}
\label{subsec:edited-summary}

For the edited summaries written by the annotators, we evaluate (1) their factual consistency; 
(2) their textual similarity with either the reference summaries or the initial outputs;
(3) other aspects of their intrinsic quality~\citep{grusky-etal-2018-newsroom, bommasani-cardie-2020-intrinsic}.

\noindent \textbf{Factual Consistency} To evaluate the factual consistency, we use two automatic metrics, DAE~\citep{goyal-durrett-2020-evaluating} and QAFactEval~\citep{fabbri-etal-2022-qafacteval}, which achieve strong performance on the XSum dataset~\citep{Tang2022UnderstandingFE}.\footnote{Automatic metric setting details are in Appendix~\ref{sec:fact}.}
The results are in Tab.~\ref{tab:fact}, showing that the human-edited summaries are more factually consistent than both the reference summaries and initial system outputs.

\paragraph{Text Similarity} For textual similarity, we compare human-edited summaries against both the reference summaries and initial system outputs in Tab.~\ref{tab:text}.
We note two observations: (1) There is a high-degree similarity between the initial system outputs and human-edited summaries, indicating that the annotators only made small changes to the initial outputs. 
(2) Compared with the initial system outputs, the human-edited summaries have lower similarity with the reference summaries, which suggests that the reference summaries and initial system outputs may share similar factual errors, leading to higher textual similarity.

\begin{table}[t!]
\centering
\small
\begin{tabular}{lcc}
\toprule
 & \textbf{DAE} & \textbf{QAFactEval}  \\
\midrule
\textbf{Reference} & 0.6176  &  1.549 \\
\textbf{System-output} & 0.6904 &  1.826 \\
\textbf{Human-edited} & \textbf{0.8975}  &  \textbf{2.540} \\
\bottomrule
\end{tabular}
\caption{Automatic factuality scores on the reference summaries, initial system outputs and human edited summaries.
The human edited summaries in \dataset dataset have significantly higher ($p<0.01$) factual consistency according to the automatic factuality metrics.}
\label{tab:fact} 
\end{table}

\begin{table}[t!]
\centering
\small
\begin{tabular}{lccc}
\toprule
 & \textbf{R1} & \textbf{R2} & \textbf{RL} \\
\midrule
\textbf{Ref. v.s. Sys.} & 48.01 & 25.54  & 40.45  \\
\textbf{Ref. v.s. Human} &  40.30 & 18.22 & 33.86 \\
\textbf{Sys. v.s. Human} & 75.79 & 66.17 & 74.89 \\
\bottomrule
\end{tabular}
\caption{Textual similarity. \textbf{R1}, \textbf{R2}, \textbf{RL} stand for the ROUGE-1/2/L F1 scores respectively. \textbf{Ref.} denotes reference summaries, \textbf{Sys.} denotes initial system outputs, and \textbf{Human} denotes the human-edited summaries.}
\label{tab:text} 
\end{table}

\begin{table}[t!]
\small
\centering
\begin{tabular}{lccc}
\toprule
 & \textbf{Coverage} & \textbf{Novelty} &  \textbf{Compression} \\
\midrule
\textbf{Ref.} & 0.633 & 0.851  & 14.82  \\
\textbf{Sys.} & 0.699 & 0.788  & 17.84  \\
\textbf{Human} & 0.787 & 0.703 & 20.61 \\
\bottomrule
\end{tabular}
\caption{Intrinsic evaluation of summary quality. 
\textit{Coverage} is negatively correlated with abstractiveness while \textit{Novelty} has a positive correlation. \textit{Compression} is the ratio of the summary length against the article length.}
\label{tab:ins} 
\end{table}

\paragraph{Intrinsic Evaluation}
We evaluate two \textit{intrinsic} summary qualities: the \textbf{\textit{compression rate}}~\citep{grusky-etal-2018-newsroom} and the \textbf{\textit{abstractiveness}}~\citep{bommasani-cardie-2020-intrinsic}.
In particular, \textit{compression rate} measures the length difference between the input text and the summary.
And to evaluate \textit{abstractiveness} we use two features, (1) Extractive Fragment Coverage~\citep{grusky-etal-2018-newsroom}, which measures the extent to which the summary can be ``copied'' from the input text, (2) Novelty, which measures the ratio of words in the summary that are not in the input text.\footnote{More details can be found in Appendix~\ref{subsec:intrinsic}.}
The statistics in Tab.~\ref{tab:ins} suggest that the human-edited summaries are less abstractive than the initial system outputs and reference summaries. 
This finding is coherent with \citet{xiao-etal-2022-datalab} which found that there exists a tradeoff between faithfulness and abstractiveness.
However, we note that the decrease of abstractiveness can result from removing non-factual information from the summary, which is the most common operation for correct \textit{extrinsic} errors, as we will show next.

\begin{figure}[t!]
    \centering
    \includegraphics[width=1\linewidth]{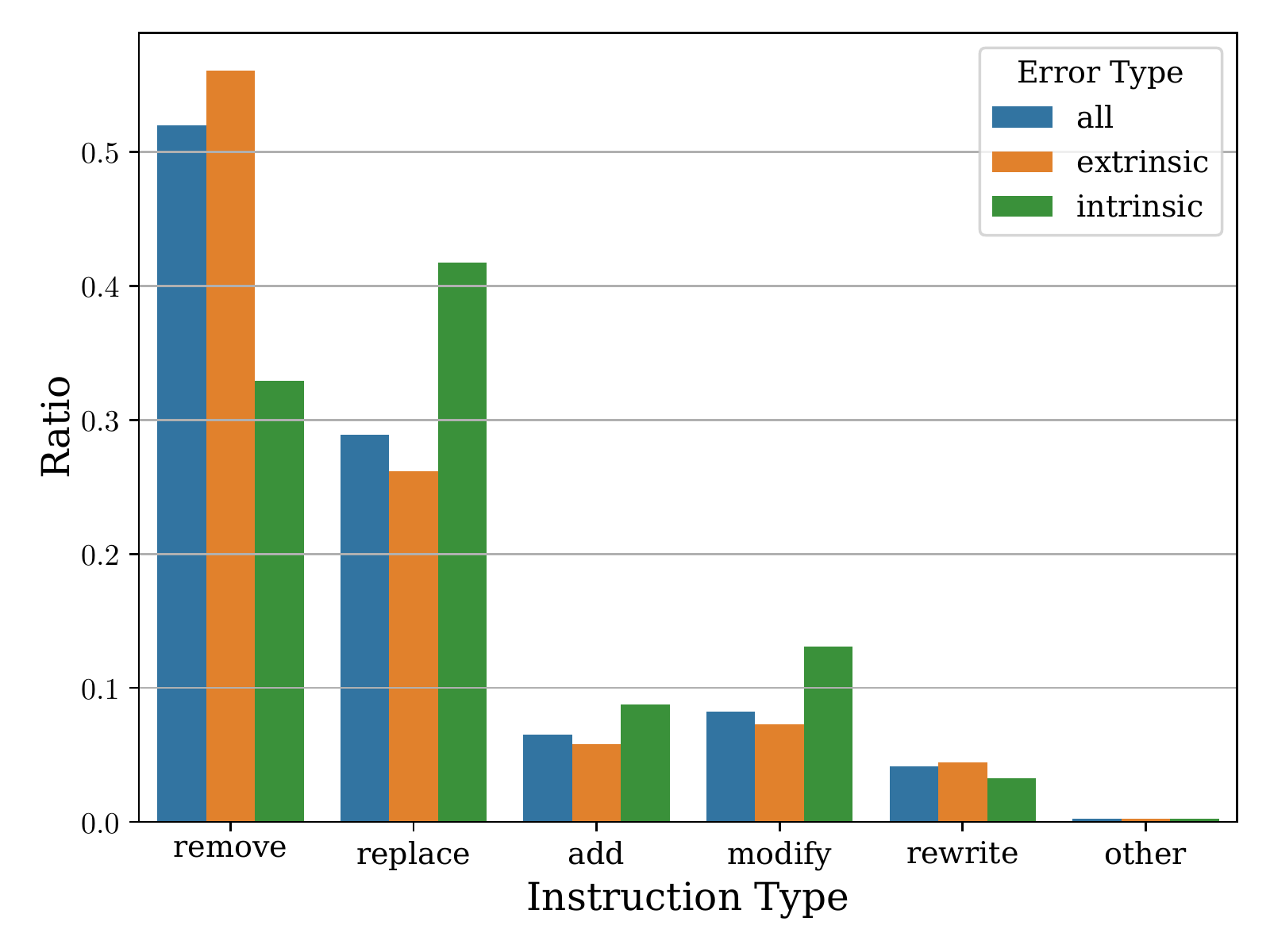}
    \caption{Distribution of six different types of instructions. \textit{removing information} and \textit{replacing information} are the most frequent types. \textit{Extrinsic} errors are more likely to be corrected by \textit{removing} while \textit{Intrinsic} errors are more likely to be corrected by \textit{replacing}.}
    \label{fig:type}
\end{figure}

\subsection{Instructions}

The annotators need to provide instructions on how to make changes to the initial system outputs to correct factual errors. 
We find that the editing can take more than one instruction and the average number of instructions is 1.52.
We show the distribution of the number of instructions in  Appendix~\ref{subsec:appendix-ins}.
As for the distribution of instruction types (Fig.~\ref{fig:type}), we found that \textbf{\textit{removing information}} and \textbf{\textit{replacing information}} to be the most frequent operations.
Interestingly, fine-grained analysis in Fig.~\ref{fig:type} shows that \textit{extrinsic} errors are more likely to be corrected by the \textit{replacing operation} while \textit{intrinsic} errors can require more diverse types of operations.

\section{Summary Editing for Improving Factual Consistency}
\label{sec:edit}

With \dataset, we propose a new NLG task: editing the initial summary based on human feedback.

\subsection{Methods}
\label{subsec:edit-method}

We formulate the summary editing task as a sequence-to-sequence  (Seq2Seq)~\citep{10.5555/2969033.2969173} problem.
Specifically, a Seq2Seq model $g$ learns a mapping from an input sequence $X$ to a target output sequence $Y$: $Y \leftarrow g(X)$.
For this specific task, the input $X$ has three components: \textit{input document}, \textit{initial system-generated summary} and \textit{human feedback}, while the target output is the \textit{human-edited summary} (Fig.~\ref{fig:intro}).
The human feedback consists of the \textit{instructions} and \textit{explanation}.
To concatenate the different components of the input sequence, a short ``\textit{prompt}'' is appended at the beginning of each component, then the entire input sequence becomes: ``Article: \textit{input document}; Candidate: \textit{initial system-generated summary}; Instruction: \textit{human instructions};  Explanation: \textit{human explanation}''.
While recent work~\citep{sanh2022multitask, bach-etal-2022-promptsource} has shown that prompt design can affect the model performance, for simplicity we use simple text snippets for the baseline models.

We instantiate the Seq2Seq model using a family of pre-trained Encoder-Decoder models, T5~\citep{JMLR:v21:20-074} and T0~\citep{sanh2022multitask}, which are widely used for transfer learning and few-shot learning where the data is scarce. 
To achieve better performance, the model is fine-tuned on the training set of \dataset using Maximum Likelihood Estimation (MLE) under the training paradigm of \textit{teacher forcing}~\citep{10.1162/neco.1989.1.2.270}.
We note that we only used the subset of data in which the initial system output contains factual errors.

\begin{table}[t!]
\small
\centering
\begin{tabular}{lccccc}
\toprule
 & \textbf{R1} & \textbf{R2} & \textbf{RL} & \textbf{DAE} & \textbf{QFE}\\
\midrule
\textbf{Sys.} & 75.98 & 66.32  & 75.05 & 0.704  &  1.837 \\
\textbf{Human.} & 100 & 100  & 100 & 0.905  &  2.550 \\
\midrule
\textbf{D+S} & 77.04 & 67.96 &  76.03 & 0.835 &  2.248 \\
\textbf{S+I} & 87.48 & 81.72 &  86.16 & 0.857 &  2.289 \\
\textbf{D+S+I} & 88.74 & 83.16 &  87.48 & 0.904 &  \textbf{2.470} \\
\textbf{D+S+E} & 81.83 & 74.10 &  80.36 & 0.899 &  2.437 \\
\textbf{D+S+I+E} & \textbf{89.22} & \textbf{83.64} &  \textbf{87.92} & \textbf{0.911} &  2.465 \\
\bottomrule
\end{tabular}
\caption{\textit{Editing} model performance (T0pp) with different variants of input. \textbf{R1}, \textbf{R2}, \textbf{RL} stand for the ROUGE-1/2/L F1 scores calculated against the human-edited summary. \textbf{DAE} is the DAE~\citep{goyal-durrett-2020-evaluating} factuality metric while \textbf{QFE} is the the QAFactEval~\citep{fabbri-etal-2022-qafacteval} metric. \textbf{Sys.} denotes initial system outputs, and \textbf{Human} denotes the human-edited summaries.
\textbf{D}, \textbf{S}, \textbf{I}, \textbf{E} stand for input \textbf{D}ocument, initial \textbf{S}ummary, human-written \textbf{I}nstructions, human-written \textbf{E}xplanation respectively.
The combinations of \textbf{D}, \textbf{S}, \textbf{I}, \textbf{E} stand for different input variants.
}
\label{tab:t0pp} 
\end{table}

\begin{table*}[t!]
\small
\centering
\extracolsep{1pt}
\begin{tabular}{@{\extracolsep{2pt}}lcccccccccccccc@{}}
\toprule
& \multicolumn{4}{c}{\textbf{T0pp}} & \multicolumn{4}{c}{\textbf{T0-3B}} & \multicolumn{4}{c}{\textbf{T5-3B}} \\
& \textbf{R1} & \textbf{R2} & \textbf{DAE} & \textbf{QFE} & \textbf{R1} & \textbf{R2} & \textbf{DAE} & \textbf{QFE} & \textbf{R1} & \textbf{R2} & \textbf{DAE} & \textbf{QFE} \\ \cmidrule{2-5} \cmidrule{6-9} \cmidrule{10-13}
\textbf{D+S} & \textit{77.04} & \textit{67.96} &  \textit{0.835} &  \textit{2.248} & 76.10 & 66.66 &  0.821 &  2.168 &  75.99 & 66.35 &  0.784 &  2.063 \\
\textbf{S+I} & \textit{87.48} & \textit{81.72} &  \textit{0.857} &  2.289  & 87.30 & 81.00 &  0.852 &  \textit{2.263} &  87.59 & 81.50 &  0.844 &  2.237 \\
\textbf{D+S+I} & \textit{88.74} & \textit{83.16} &  \textit{0.904} &  2.470  & 88.36 & 82.16 &  0.894 &  \textit{2.489} & 86.42 & 80.56 &  0.876 &  2.411 \\
\textbf{D+S+E} & \textit{81.83} & \textit{74.10} &   \textit{0.899} &  2.437   & 79.85 & 71.41 &  0.902 &  \textbf{\textit{2.510}} &  79.09 & 71.20 &  0.877 &  2.373 \\
\textbf{D+S+I+E} & \textbf{\textit{89.22}} & \textbf{\textit{83.64}} &  \textbf{\textit{0.911}} &  2.465 & 88.69 & 82.44 &   0.899 &  \textit{2.477} & 87.03 & 80.77 &   0.865 &  2.375 \\
\bottomrule
\end{tabular}
\caption{Performance with different variants of the \textit{Editing} model. \textbf{R1}, \textbf{R2} stand for the ROUGE-1/2 F1 scores. \textbf{DAE} is from \citet{goyal-durrett-2020-evaluating} while \textbf{QFE} is from \citet{fabbri-etal-2022-qafacteval}.
\textbf{D}, \textbf{S}, \textbf{I}, \textbf{E} stand for input \textbf{D}ocument, initial \textbf{S}ummary, human-written \textbf{I}nstructions, human-written \textbf{E}xplanation respectively, of which the combinations stand for different input variants.
The best results with each input variant are \textit{italicized}.
}
\vspace{-3mm}
\label{tab:edit} 
\end{table*}

\subsection{Experiments}
\label{subsec:edit-exp}

\noindent \textbf{Implementation Details}
To initialize the \textit{Editing} model, we use T5-3B
and two variants of T0 models, T0-3B
and T0pp.\footnote{T5-3B (\url{https://huggingface.co/t5-3b}), T0-3B (\url{https://huggingface.co/bigscience/T0_3B}), and T0pp (\url{https://huggingface.co/bigscience/T0pp}) have around 3, 3, and 11 billion parameters respectively.}
We compare the model performance with different variants of input (e.g., with or without the human-written explanation).
To evaluate the quality of the model output, we focus on two aspects: \textit{textual similarity} with the human-edited summary, as evaluated by ROUGE~\citep{lin-2004-rouge}, and \textit{factual consistency} with the input document, as evaluated by \textbf{DAE}~\citep{goyal-durrett-2020-evaluating} and QAFactEval (\textbf{QFE})~\citep{fabbri-etal-2022-qafacteval}.
The checkpoints are selected based on their performance on the validation set.

\noindent \textbf{Experimental Results}
Tab.~\ref{tab:t0pp} shows the performance of fine-tuned T0pp with different input variants.
We note the following observations:
(1) Compared with the initial system-generated summaries, the \textit{Editing} model is able to generate summaries more similar to the human-edited summaries and more factually consistent with the input document.
(2) Both the human-written instructions and explanation can provide meaningful guidance to the \textit{Editing} model, and the model with both of them as input (the \textbf{D+S+I+E} variant) achieves the best performance.
(3) Without the input document, the \textit{Editing} model (the \textbf{S+I} variant) can still improve the initial system-generated summaries by following the instructions.
However, taking the input document as part of the input helps the model (the \textbf{D+S+I} variant) to achieve better performance, especially for better factual consistency.

In Tab.~\ref{tab:edit}, we compare the performance of T0pp, T0-3B and T5-3B with different kinds of inputs. 
We found that (1) the findings on T0pp (Tab.~\ref{tab:t0pp}) are generalizable to T0-3B and T5-3B with few exceptions. 
(2) T0pp outperforms T0-3B across different input variants according to different automatic metrics except for the QAFactEval metric. 
(3) T0-3B generally outperforms T5-3B, likely thanks to the pre-training of T0 which is designed for performing zero-shot learning with instructions.

\noindent \textbf{Human Evaluation}
We conduct a human evaluation on the quality of model-edited summaries.
We ask the annotators two questions: (1) Are the generated summaries more factually consistent than the original summaries (yes/no); (2) Do the generated summaries follow the instructions (yes/partly/no).
We randomly sampled 100 examples from the test set, and have each generated summary annotated by three MTurk annotators.
The generated summaries are from the trained checkpoint of T0pp with the input containing input document, initial system-generated summaries, and human-written instructions. 
Under major voting (with ties ignored), we found that 97\% of model-edited summaries are more factually consistent than the original system outputs, and 91\% of them follow the provided human-written instructions.

\subsection{Case Study of LLM Summary Editing}
\label{subsec:llm-edit}

As a case study, we evaluate the zero-shot learning ability of GPT-3.5\footnote{OpenAI's \texttt{gpt-3.5-turbo-0301}: \url{https://platform.openai.com/docs/models/gpt-3-5}.} for summary editing. 
We apply it to two settings, (1) editing without instructions and (2) editing by following instructions, in a zero-shot learning manner.\footnote{The prompts we used can be found in Appendix~\ref{appendix:llm-edit}.}
The results in Tab.~\ref{tab:llm-edit} show that (1) GPT-3.5 is able to leverage the editing instructions; (2) Compared with the fine-tuned model (T0pp), GPT-3.5 can generate edited summaries with higher factual consistency but it is worse at maintaining the content similarity with the original summary, which suggests that it still struggles with \textit{controllable} text generation. 

\begin{table}[t!]
\small
\centering
\begin{tabular}{lccccc}
\toprule
\textbf{Model} & \textbf{Input} & \textbf{R1} & \textbf{R2} & \textbf{DAE} & \textbf{QFE}\\
\midrule
\textbf{Sys.} & - & 75.98 & 66.32     & 0.704 & 1.837 \\
\textbf{Human.} & - & 100 & 100     &  0.905 & 2.550 \\
\midrule
\textbf{T0pp} & \textbf{D+S}  & 77.04 & 67.96 & 0.835 &  2.248 \\
\textbf{T0pp} & \textbf{D+S+I} &  88.74 & 83.16 &  0.904 &  2.470 \\
\midrule
\textbf{GPT-3.5} & \textbf{D+S} & 36.75 &      21.98     & 0.892 &  2.351 \\
\textbf{GPT-3.5} & \textbf{D+S+I} &  72.22 &       60.53 &   0.910 & 2.651 \\
\bottomrule
\end{tabular}
\caption{Performance of GPT-3.5 for summary editing. \textbf{R1}, \textbf{R2} stand for the ROUGE-1/2 F1 scores calculated against the human-edited summary. 
\textbf{Sys.} denotes initial system outputs, and \textbf{Human} denotes the human-edited summaries.
\textbf{D}, \textbf{S}, \textbf{I} stand for input \textbf{D}ocument, initial \textbf{S}ummary, human-written \textbf{I}nstructions.  
}
\vspace{-3mm}
\label{tab:llm-edit} 
\end{table}

\section{Generating Feedback for Improving Factual Consistency}

We investigate if it is possible to train a model to generate feedback from a given document and summary pair to correct factual errors, and we name the subsequent model as a \textit{Critic} model. 

\subsection{Methods}
\label{subsec:ins-method}

Similarly to \S\ref{subsec:edit-method}, we formulate the \textit{Critic} model as a Seq2Seq model.
The input sequence is a concatenation of the \textit{input document} and the \textit{initial system-generated summary} while the target output is the \textit{human-written instructions} (Fig.~\ref{fig:intro}).\footnote{While it is also possible to require the \textit{Critic} model to generate the explanation, we choose human-written instructions as the target output as it works better at helping the \textit{Editing} model to improve the initial outputs (\S\ref{subsec:edit-exp}).}   
We use T0 as the startpoint to fine-tune the \textit{Critic} model with MLE training on the subset of \dataset in which the initial summary contains factual errors. 

\subsection{Experiments}

\begin{table}[t!]
\centering
\small
\begin{tabular}{lccc}
\toprule
& \textbf{Rouge1} & \textbf{Rouge2} & \textbf{RougeL}\\
\midrule
\textbf{T0pp} & 52.55 & 37.41 &  51.00 \\
\textbf{T0-3B} & 51.70 & 36.56 &  50.33  \\
\bottomrule
\end{tabular}
\caption{\textit{Critic} model performance with respect to the textual similarity between the system output and human-written instructions.
}
\label{tab:critic} 
\end{table}

\begin{table}[t!]
\small
\centering
\begin{tabular}{lccccc}
\toprule
\textbf{Method} & \textbf{Critic} & \textbf{R1} & \textbf{R2} & \textbf{DAE} & \textbf{QFE}\\
\midrule
\textbf{Sys.} & - & 75.98 & 66.32     & 0.704 & 1.837 \\
\textbf{Human.} & - & 100 & 100     &  0.905 & 2.550 \\
\midrule
\textbf{D+S} & - & 77.04 & 67.96 & 0.835 &  2.248 \\
\textbf{D+S+I} & - & 88.74 & 83.16 &  0.904 &  2.470 \\
\midrule
\textbf{D+S+$\textrm{I}^*$} & T0pp &  75.10  &  65.15 &  0.859 & 2.296 \\
\textbf{D+S+$\textrm{I}^*$} & T0-3B & 73.01   & 62.15 & 0.859 & 2.278 \\
\bottomrule
\end{tabular}
\caption{\textit{Editing} model performance (T0pp) with the instructions generated by the \textit{Critic} model. \textbf{R1}, \textbf{R2} stand for the ROUGE-1/2 F1 scores calculated against the human-edited summary. %
\textbf{Sys.} denotes initial system outputs, and \textbf{Human} denotes the human-edited summaries.
\textbf{D}, \textbf{S}, \textbf{I} stand for input \textbf{D}ocument, initial \textbf{S}ummary, human-written \textbf{I}nstructions.  
\textbf{$\textrm{I}^*$} stands for the instructions generated by the \textit{Critic} model.
}
\label{tab:critic-edit} 
\end{table}

\noindent\textbf{Experimental Results}
Tab.~\ref{tab:critic} shows the textual similarity between the instructions generated by the \textit{Critic} model and the human-written instructions.
To have a more intuitive understanding of the model performance, in Tab.~\ref{tab:critic-edit} we evaluate the performance of the \textit{Editing} model with the instructions generated by the \textit{Critic} model.
We found that 
(1) While the generated instructions cannot work as well as the human-written instructions, they are helpful to the \textit{Editing} model to improve the factual consistency of the initial system-generated summaries.
(2) Compared with the \textit{Editing} model that only takes the input document and initial summary as input (\textbf{D+S}), the \textit{Editing} model (\textbf{D+S+I*}) that also uses the generated instructions achieves better performance with respect to factual consistency, but its outputs have lower textual similarity with the human-edited summaries. 
It indicates that the \textit{Critic} model can generate useful instructions.
Meanwhile, the lower textual similarity may result from the fact that there can be more than one way to correct the factual errors,\footnote{For example, one may choose to \textit{remove} a factual error or \textit{replace} the error with factual information when appropriate.} and the \textit{Critic} model can generate instructions for a way of correction different from the human-edited summary. 

\noindent\textbf{Human Evaluation} To further evaluate the quality of model-generated instructions, we ask human annotators two questions: (1) Are the generated instructions equivalent to human-written instructions (yes/partly/no); (2) Are the generated instructions useful for correcting the factual errors (yes/partly/no).
Similar to \S\ref{subsec:edit-exp}, we randomly sampled 100 examples from the test set, and have each generated instruction annotated by three MTurk annotators.
For the first question, we found that the annotators think 24\% of the generated instructions are \textit{exactly} equivalent to the human-written instructions while 45\% of them are \textit{partly} equivalent.
For the second question, we found that 39\% of the generated instructions are useful for correcting factual errors while 31\% of them are partly useful.
As a result, we found that it is easier for the \textit{Critic} model to generate useful instructions than generating instructions that are equivalent to human-written instructions. 
We hypothesize this is because there can be more than one way to edit the initial summary therefore the human-written instructions represent only one acceptable solution.

\subsection{Case Study of LLM Critic}
\label{subsec:llm-critic}
\begin{table}[t!]
\small
\centering
\begin{tabular}{lccccc}
\toprule
\textbf{Method} & \textbf{Critic} & \textbf{R1} & \textbf{R2} & \textbf{DAE} & \textbf{QFE}\\
\midrule
\textbf{Sys.} & - & 75.98 & 66.32     & 0.704 & 1.837 \\
\textbf{Human.} & - & 100 & 100     &  0.905 & 2.550 \\
\textbf{D+S+$\textrm{I}^*$} & T0pp &  75.10  &  65.15 &  0.859 & 2.296 \\
\midrule
\textbf{D+S+$\textrm{I}^*$} & GPT-3.5 &   60.48 &   48.18 &  0.868 & 2.566 \\
\textbf{D+S+$\textrm{I}^*$} & GPT-4 & 63.60 &       51.15  & 0.860 & 2.604 \\
\bottomrule
\end{tabular}
\caption{Case study of instruction generation with LLMs.
Instructions generated by the \textit{Critic} models are used to instruct the \textit{Editing} model.
}
\label{tab:critic-llm} 
\end{table}

As a case study, we evaluate the zero-shot learning ability of GPT-3.5 and GPT-4\footnote{OpenAI's \texttt{gpt-4-0314}: \url{https://platform.openai.com/docs/models/gpt-4}.} for instruction generation.
The results in Tab.~\ref{tab:critic-llm} show that, compared with fine-tuned models, instructions generated by both GPT-3.5 and GPT-4 lead the editing model to generate summaries that are more factual but less similar to the original summaries.
This finding shows a similar trend as in \S\ref{subsec:llm-edit}, that LLMs in a zero-shot learning setting lack the ability of \textit{controllable} text generation.
For example, GPT-3.5 responded with ``No editing instructions needed'' 23.9\% of the time, despite being directly instructed to edit a factually inconsistent summary.\footnote{Prompts and more details are in Appendix~\ref{appendix:llm-critic}.}

\section{Summary Editor with Feedback Generation and Editing}

We define the third NLG task as to predict both the \textbf{\textit{human feedback}} and the \textbf{\textit{edited summary}} given the input document and initial system-generated summary.
We name this model the \textbf{\textit{Editor}} model because it needs to both evaluate the initial summary and make edits according to its own assessments.

\begin{table}[t!]
\small
\centering
\begin{tabular}{lccccc}
\toprule
 & \textbf{Method}  & \textbf{R1} & \textbf{R2} & \textbf{DAE} & \textbf{QFE}\\
\midrule
& Sys. & 75.98 & 66.32     & 0.704 & 1.837 \\
& Human. & 100 & 100     &  0.905 & 2.550 \\
\midrule
\multirow{3}{*}{\textbf{T0pp}}& Editing & 77.04 & 67.96 & 0.835 &  2.248 \\
& $\textrm{Editor}_{I}$  & 78.01 & \textbf{69.01} & 0.804  &  2.108 \\
& $\textrm{Editor}_{E}$  & \textbf{78.46} & 68.70 & \textbf{0.867} & \textbf{2.309} \\
\midrule
\multirow{3}{*}{\textbf{T0-3B}}& Editing & 76.10 & 66.66 &  0.821 &  2.168 \\
& $\textrm{Editor}_{I}$  & \textbf{77.40} & \textbf{68.29} & 0.808  & 2.112  \\
& $\textrm{Editor}_{E}$  & 77.27 & 67.92 & \textbf{0.838}  &  \textbf{2.241} \\
\midrule
\multirow{3}{*}{\textbf{T5-3B}}& Editing & 75.99 & 66.35 &  0.784 &  2.063 \\
& $\textrm{Editor}_{I}$  & \textbf{77.06} &  \textbf{67.86}  & \textbf{0.804} & 2.106\\
& $\textrm{Editor}_{E}$  & 76.82 & 67.42 & 0.796 &  \textbf{2.114} \\
\bottomrule
\end{tabular}
\caption{\textit{Editor} model performance.  \textbf{R1}, \textbf{R2} stand for the ROUGE-1/2 F1 scores calculated against the human-edited summary. 
\textbf{Sys.} denotes initial system outputs, and \textbf{Human} denotes the human-edited summaries.
\textit{Editing} is the model in \S\ref{sec:edit} with only the input document and initial system-generated summary as input.
$\textrm{Editor}_{I}$ is the \textit{Editor} model that generates both the instructions and edited summary, while $\textrm{Editor}_{E}$ is the one that generates both the explanation and edited summary.
}
\label{tab:editor} 
\end{table}

\subsection{Correcting Known Factual Errors}

Similar to \S\ref{subsec:edit-method} and \S\ref{subsec:ins-method}, we fine-tuned the pre-trained T0 and T5 models for our experiments. 
The two parts of the target output, the human feedback, and the edited summary are indicated by textual tags as specified in \S\ref{subsec:edit-method}.
We investigate two specific scenarios: (1) generating both the \textit{instructions} and the edited summary; (2) generating both the \textit{explanation} and the edited summary.

We present the experimental results in Tab.~\ref{tab:editor}.
Compared with the \textit{Editing} model that takes only the input document and the initial system-generated summary as the input, the \textit{Editor} models have better performance in textual similarity, and the one that generates the explanations also achieves higher factual consistency. 
The results suggest that learning to predict related information of a target generation task can be beneficial to the performance of language generation models, echoing the recent findings in chain-of-thought prompting~\citep{Wei2022ChainOT, Huang2022ChainOE, Jung2022MaieuticPL}.

\begin{table}[t!]
\small
\centering
\begin{tabular}{lccccc}
\toprule
\textbf{System} & \textbf{R1} & \textbf{R2} & \textbf{RL} & \textbf{DAE} & \textbf{QFE}\\
\midrule
\textbf{Pegasus} & 47.35 &    24.61 &   39.59 &  0.763 & 2.029 \\
\textbf{Human} &  41.94 &    19.49 &   34.97      &  0.905 & 2.550 \\
\midrule
\textbf{CCGS} &  45.11  &  21.06 & 36.60 & 0.760  & 1.847 \\
\textbf{CLIFF} &    \textbf{46.40} &    \textbf{23.38} &    \textbf{38.38} & 0.780 & 2.068 \\
\textbf{ReDRESS} & 43.50 &   19.77 &   35.28 & 0.830 &   2.065 \\
\textbf{FactPegasus} &    38.95 &    15.99 &   31.68 &        \textbf{0.882} &  1.941 \\
\textbf{CompEdit} &    42.69 &   19.06 &     34.73 &       0.850 & 2.113 \\
\textbf{Editor} &  45.14 &    22.27 &    37.89  &     0.833 & \textbf{2.250} \\
\bottomrule
\end{tabular}
\caption{\textit{Editor} model performance (T0-3B) on the \textit{entire} \dataset test set. \textbf{R1}, \textbf{R2}, \textbf{RL} stand for the ROUGE-1/2/L F1 scores calculated against the \textit{reference} summary. \textbf{DAE} is the DAE~\citep{goyal-durrett-2020-evaluating} factuality metric while \textbf{QFE} is QAFactEval~\citep{fabbri-etal-2022-qafacteval}. 
The initial system outputs are from \textbf{Pegasus}, and \textbf{Human} are the human-edited summaries.
}
\label{tab:editor-comp} 
\end{table}

\subsection{Detecting and Correcting Factual Errors}

\label{subsec:6-2}
While in the previous experiments the models are trained to edit the initial system outputs with known factual errors, the \textit{Editor} model can also be used on \textit{arbitrary} system outputs where it is required to edit the initial output \textit{only} when it identifies factual errors in it.
To this end, we use the entire \dataset in this experiment with the following modifications to the target output:
(1) the target summary is set to the original system output when it contains no factual errors, and to the human-edited summary otherwise;
(2) only explanations are used as part of the target output because it is always available. 

We fine-tune T0-3B in this experiment and compare its results with several recently introduced summarization systems that are specifically trained to improve the summary factual consistency:
(1) CCGS~\citep{chen-etal-2021-improving},
(2) CLIFF~\citep{cao-wang-2021-cliff},
(3) ReDRESS~\citep{adams2022learning}, (4) FactPegasus~\citep{wan-bansal-2022-factpegasus}, (5) CompEdit~\citep{Fabbri2022ImprovingFC}.
More details about these systems can be found in Appendix~\ref{appendix:sys}.

The results in Tab.~\ref{tab:editor-comp} show that the \textit{Editor} can achieve competitive performance compared with the baseline systems and yield a balanced performance between the \textit{content similarity} with the reference summary and the \textit{factuality consistency}.
Since the \textit{Editor} model is trained on much fewer data than the other systems, its strong performance indicates the effectiveness of utilizing human demonstrations and feedback for improving factual consistency.

\section{Related Work}
\noindent \textbf{Factual Consistency in Text Summarization}
Factual consistency is an important quality of text summarization systems~\citep{kryscinski-etal-2020-evaluating, maynez-etal-2020-faithfulness, fabbri-etal-2021-summeval}.
Related work has proposed various methods of improving the factual consistency of summaries by (1) training abstractive summarization models with factuality metrics~\citep{goyal-durrett-2021-annotating, cao-etal-2022-hallucinated}, (2) introducing new training objectives and multi-task learning for model training~\citep{cao-wang-2021-cliff, zhu-etal-2021-enhancing, aralikatte-etal-2021-focus,  zhang-etal-2022-improving-faithfulness, xu-zhao-2022-jointly}, (3) post-editing or re-ranking the initially generated summaries to improve the factual consistency~\citep{cao-etal-2020-factual, chen-etal-2021-improving, https://doi.org/10.48550/arxiv.2210.12378, Fabbri2022ImprovingFC}, (4) designing factuality-aware pre-training~\citep{wan-bansal-2022-factpegasus}.

To facilitate the evaluation of summarization models and automatic factuality metrics that evaluate the factual consistency of summaries, various benchmark datasets have been collected by the related work~\citep{kryscinski-etal-2020-evaluating, wang-etal-2020-asking, huang-etal-2020-achieved, fabbri-etal-2021-summeval, goyal-durrett-2021-annotating, pagnoni-etal-2021-understanding}.
In these benchmarks, system-generated summaries are evaluated by human annotators with either numerical quality scores, binary labels, or binary labels with fine-grained error taxonomies. 
In contrast, our dataset contains more detailed human feedback with \textit{natural language descriptions} and provides error-free, human-edited summaries. 

\noindent\textbf{Neural Text Editing}
Neural text editing models~\citep{malmi-etal-2022-text} are suitable for application scenarios where there is a significant textual overlap between the input and output sequences~\citep{awasthi-etal-2019-parallel,malmi-etal-2019-encode, stahlberg-kumar-2020-seq2edits, mallinson-etal-2020-felix, reid-zhong-2021-lewis, https://doi.org/10.48550/arxiv.2205.12209}, such as grammatical error correction, text simplification, and text style transfer.
Instead of autoregressive generation, text editing can also be achieved by predicting and performing edit operations~\citep{stahlberg-kumar-2020-seq2edits, mallinson-etal-2020-felix} or through non-autoregressive text generation~\citep{10.5555/3454287.3455290, agrawal-carpuat-2022-imitation}.
Unlike most of the related work, we propose a text editing task that requires the editing models to follow the editing instructions. 
\citet{faltings-etal-2021-text} introduces a similar dataset as ours containing single-sentence edits and the associated natural language commands crawled from Wikipedia. 
However, our dataset is different from theirs as we define a specific target quality, summary factual consistency, for the text edits and instructions.

\noindent\textbf{Improving Neural Models through Human Feedback}
Leveraging human feedback to improve neural models has become a recent research focus.
InstructGPT3~\cite{Ouyang2022TrainingLM} use human feedback to improve initial predictions from a GPT3 model for better user preference alignments.
\citet{Madaan21} propose the interactive MERCURIE system, where users interactively correct the explanations generated by a reasoning system. 
In \citet{BlenderBotFITS}, a generic chatbot is continuously trained using various forms of human feedback including natural language comments.
\citet{PEERCOLLABLM} propose a collaborative language model, PEER, which imitates a draft writing process and interactively refines a language generation task through human feedback. 
For text summarization, prior  works~\citep{stiennon2020learning, https://doi.org/10.48550/arxiv.2109.10862, nguyen-etal-2022-make, LMLanguageFeedback} have studied training summarization models through human feedback in the form of numerical scores of summary quality and thus different from natural language feedback used in our work.

\section{Conclusions}

Using summary factual consistency as a target quality, we study improving text generation with human demonstrations and feedback.
We demonstrate the usages of human feedback in three proposed NLG tasks using the collected dataset, \dataset, and show that human feedback can be used to improve summary factual consistency. 
We believe that our proposed tasks can be extended to other important text qualities beyond factual consistency, and utilizing natural language feedback for improving text generation can be a promising path for future work.  

\section*{Acknowledgements}
We thank the anonymous reviewers for their valuable feedback and helpful suggestions.

\clearpage
\section*{Limitations}

The annotation task we proposed in this work, i.e., detecting factual errors in summaries and providing human demonstrations and feedback for correcting the identified errors, can be complicated and time-consuming.
During our recruiting phase for MTurk annotators, we found that the ratio of annotators who were qualified after finishing the qualification test was relatively low.
Therefore, it can be difficult to scale up the annotated dataset given the time and budget limitations.
As a result, our dataset is of a relatively small scale and we only used one summarization dataset (XSum) and one base summarization model (Pegasus).

In this work, we view summary factual consistency as an example of user-expected quality to study leveraging natural language feedback for aligning system outputs with user preferences.
However, user preferences can be diverse and personal and some user-expected output quality will be less well-defined and objective than summary factual consistency, which further increases the difficulty and ambiguity of data annotation and model evaluation. 
Therefore, it can be challenging to directly apply the methods we proposed in this work to such subjective quality aspects, and we leave it for future work to explore generalizing our methods to more diverse user expectations and preferences.

\bibliography{anthology,custom}
\bibliographystyle{acl_natbib}

\appendix

\section{Data Collection Details}

We used the XSum dataset for our data collection.
It is released under the Apache 2 license and contains news articles written in English.

\subsection{Instruction Templates}
\label{subsec:ins-temp}
We provide six templates for the annotators corresponding to different operations: \textit{Remove}, \textit{Add}, \textit{Replace}, \textit{Modify}, \textit{Rewrite}, \textit{Others}:

\noindent (1) \textbf{\textit{Remove}} the information about \_\_ from the summary.

\noindent (2) \textbf{\textit{Add}} the information about \_\_ to the summary.

\noindent (3) \textbf{\textit{Replace}} the information about \_\_  \textit{with} the information about \_\_.

\noindent (4) \textbf{\textit{Modify}} the information about \_\_ in the summary.

\noindent (5) \textbf{\textit{Rewrite}} the summary \textit{entirely} by \_\_.

\noindent (6) \textbf{\textit{Other}} instructions: \_\_.

We note that sometimes it takes more than one instruction to edit the original summary.

\subsection{Annotated Examples}
\label{subsec:examples}
We provide the following annotated examples.

\noindent \textbf{Example 1}

\textbf{Original Summary:}
A Wirral biscuit factory is to close with the loss of 342 jobs.

\textbf{Explanation:}
Location is in Moreton (Morton?), not Wirral, and 342 families will be affected but that technically doesn't translate to 342 jobs.

\textbf{Instruction:}
Replace the information about Wirral with the information about Moreton. Replace the information about loss of 342 jobs with the information about affect 342 families.

\textbf{Edited Summary:}
A Moreton biscuit factory is to close, affecting 342 families.

\noindent \textbf{Example 2}

\textbf{Original Summary:}
Two teenage girls have appeared at Teesside Crown Court accused of murdering a woman in Middlesbrough.

\textbf{Explanation:}
The Teesside Crown Court was not mentioned by name, only the Youth court. The woman was found in Stephen Street and not Midlesbrough.

\textbf{Instruction:}
Replace the information about Middlesbrough with the information about Stephen Street. Replace the information about Teesside Crown Court with the information about Teesside Youth Court.

\textbf{Edited Summary:}
Two teenage girls have appeared at Teesside Youth Court accused of murdering a woman in Stephen Street.

\noindent \textbf{Example 3}

\textbf{Original Summary:}
Michael O'Halloran believes St Johnstone manager Tommy Wright will get the best out of him following his release by Rangers.

\textbf{Explanation:}
the first name info in summary is not found in the source text. St Johnstone info is also not mentioned in the source.

\textbf{Instruction:}
Remove the information about the first names of both people from the summary. Remove the information about Wright being the St Johnstone manager from the summary.

\textbf{Edited Summary:}
O'Halloran believes Wright will get the best out of him following his release by Rangers.

\noindent \textbf{Example 4}

\textbf{Original Summary:}
Aberdeen's Royal Concert Hall is to sell off hundreds of items of memorabilia as part of building work.

\textbf{Explanation:}
The source text doesn't state the name of the concert hall.

\textbf{Instruction:}
Replace the information about Aberdeen's Royal Concert Hall with the information about Aberdeen Performing Arts.

\textbf{Edited Summary:}
Aberdeen Performing Arts is to sell off hundreds of items of memorabilia as part of building work.

\noindent \textbf{Example 5}

\textbf{Original Summary:}
Lancashire County Council's decision to stop composting waste has been criticised as ``catastrophic''.

\textbf{Explanation:}
The original summary strongly implies that the decision to stop composting was ``catastrophic'' but the original text strongly implies that the composting program itself was a catastrophic failure versus stopping the program.

\textbf{Instruction:}
Replace the information about the stoppage of the composting program being catastrophic with the information about how the composting program was a catastrophic failure.

\textbf{Edited Summary:}
Lancashire County Council's decision to stop composting waste shows the program has been a ``catastrophic'' failure.

\noindent \textbf{Example 6}

\textbf{Original Summary:}
Alex Goode and Ollie Devoto have been called up to England's Six Nations training squad.

\textbf{Explanation:}
The source text does not mention the name of England's training squad.

\textbf{Instruction:}
Remove the information about the name of England's training squad from the summary.

\textbf{Edited Summary:}
Alex Goode and Ollie Devoto have been called up to England's training squad.

\section{Factuality Metrics Setting}
\label{sec:fact}
We use two factuality metrics, DAE~\citep{goyal-durrett-2020-evaluating} and QAFactEval~\citep{fabbri-etal-2022-qafacteval}, in our experiments. 
For DAE, we transfer its \textit{dependency-level} predictions into \textit{summary-level} scores.
Specifically, following the notation of \citet{goyal-durrett-2021-annotating}, we use $d(S)$ to denote the dependency-parse of a summary $S$.   
For each arc $a$ in $d(S)$, the DAE metric predicts a label $y_{a}$ representing if this dependency is entailed by the input document ($y_{a} = 1$ means entailment.)
Then, we define a summary-level factuality score $f_{S}$ using the predictions:
\begin{equation}
    f_{S} = \frac{\sum_{a}y_{a}}{|d(S)|}.
\end{equation}
For QAFactEval, we directly use its prediction scores since they are summary-level scores.

\section{\dataset Analysis}

\subsection{Intrinsic Summary Evaluation}
\label{subsec:intrinsic}
In \S\ref{subsec:edited-summary}, we use three metrics to evaluate the summary's intrinsic quality. 

\noindent (1) Compression rate~\citep{grusky-etal-2018-newsroom} is defined as the ratio of the number of words in the input document $D$ and in the summary $S$:
\begin{equation}
    \textrm{COMPRESSION}(D, S) = \frac{|D|}{|S|}.
\end{equation}

\noindent (2) Extractive Fragment Coverage~\citep{grusky-etal-2018-newsroom} is defined with \textit{extractive fragments}~\citep{grusky-etal-2018-newsroom}, $F(D, S)$, which is a set of word sequences shared between the input document $D$ and the summary $S$.
Then, the Extractive Fragment Coverage is defined as
\begin{equation}
\textrm{COVERAGE}(D, S) = \frac{1}{|S|}\sum_{f \in F(D, S)} |f|.
\end{equation}

\noindent (3) We define a word-level novelty between the input document $D$ and the summary $S$ as
\begin{equation}
    \textrm{NOVELTY} = 1 - \frac{|D \cap S|}{|S|}.
\end{equation}

\subsection{Human-written Instructions}
\label{subsec:appendix-ins}
\begin{figure}[t!]
    \centering
    \includegraphics[width=0.7\linewidth]{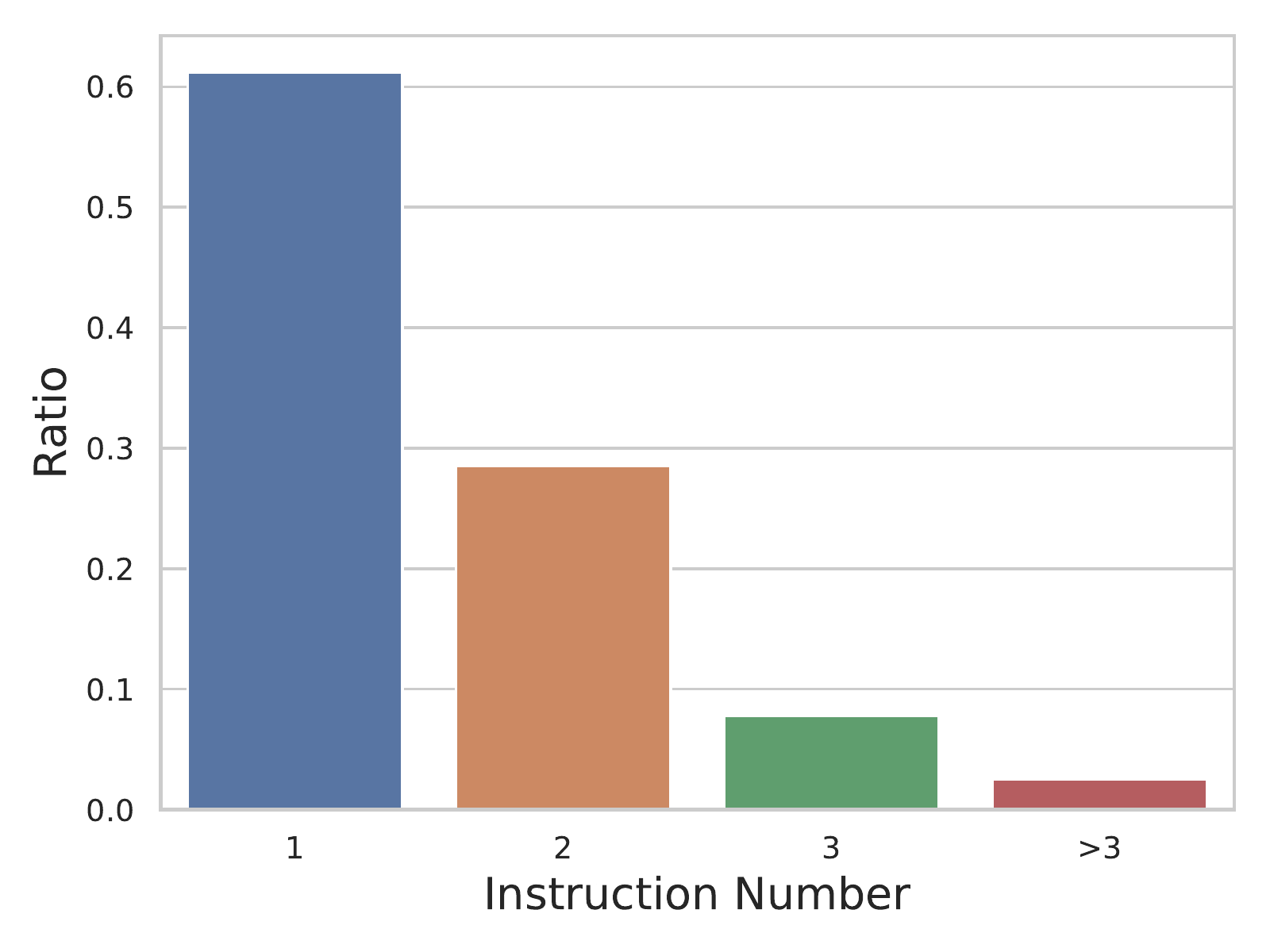}
    \caption{Number of instructions to correct a factually inconsistent system-generated summary.}
    \label{fig:num}
\end{figure}

We find that it can take more than one instruction to perform the summary editing.
Fig.\ref{fig:num} shows the distribution of the number of instructions.

\section{Experimental Details}

We use T5-3B
and two variants of T0 models, T0-3B
and T0pp in our experiments.\footnote{T5-3B (\url{https://huggingface.co/t5-3b}), T0-3B (\url{https://huggingface.co/bigscience/T0_3B}), and T0pp (\url{https://huggingface.co/bigscience/T0pp}) have around 3, 3, and 11 billion parameters respectively.}
For the 3B models, it takes one 40GB GPU to train the model, and the training time is around 8 hours.
For the 11B models, it takes eight 32GB GPUs to train the model, and the training time is around 20 hours.
All the experiments converged in 50 epochs.

\subsection{Baseline Summarization Systems}
\label{appendix:sys}
In \S\ref{subsec:6-2}, we compare the performance of \textit{\textbf{Editor}} model with the following summarization systems:

\noindent (1) CCGS~\citep{chen-etal-2021-improving}, which is based on contrastive candidate generation and selection.

\noindent (2) CLIFF~\citep{cao-wang-2021-cliff}, which is trained with contrasting learning and synthetically generated contrastive examples. 

\noindent (3) ReDRESS~\citep{adams2022learning}, which is a summary post-editor that learns to remove factual errors through contrastive learning. 

\noindent (4) FactPegasus~\citep{wan-bansal-2022-factpegasus}, which is pre-trained and fine-tuned with factual-consistency-aware training objectives.

\noindent (5) CompEdit~\citep{Fabbri2022ImprovingFC}, which is a compression-based post-editing model that removes the non-factual entities from the original summary by performing summary compression.

\subsection{Setting of LLM Case Study for Summary Editing}
\label{appendix:llm-edit}
We use GPT-3.5 for the summary editing experiment with LLMs. 
To ensure stable results, we set the sampling temperature to 0.
The prompt for summary editing \textit{without} instructions is as follows:
\begin{quote}
    You will be given an article and a summary of the article, which is not factually consistent with the article. That is, the summary contains information that is not supported by the article.
    
Your task is to edit the summary to make it factually consistent with the article. The correction should preserve most of the summary and only adapt it. Please only make the necessary changes to the summary. However, if you find all the information in the summary is not correct, please write a new summary of the entire article instead. 

The edit operations are: 1. Remove Information, 2. Add Information, 3. Replace Information, 4. Modify Information, 5. Rewrite Summary 6. Others.

The summary should contain only one sentence. Please keep the style of the summary unchanged, and the length of the summary should be similar before and after your edits.

Article: \{\{Article\}\}

Summary: \{\{Summary\}\}

Please edit the summary accordingly:
\end{quote}

The prompt for summary editing \textit{with} instructions is as follows:
\begin{quote}
    You will be given an article and a summary of the article, which is not factually consistent with the article. That is, the summary contains information that is not supported by the article.
    
You will be given instructions about how to edit the summary to make it factually consistent with the article. Your task is to follow the instructions and edit the summary accordingly. The correction should preserve most of the summary and only adapt it. Please only make necessary changes to the summary, and keep the summary length close to the original length. The summary should contain only one sentence.

Article: \{\{Article\}\}

Summary: \{\{Summary\}\}

Instructions: \{\{Instruction\}\}

Please edit the summary accordingly:
\end{quote}

\subsection{Setting of LLM Case Study for Instruction Generation}
\label{appendix:llm-critic}

The prompt we used for instruction generation is as follows:
\begin{quote}
    You will be given an article and a summary of the article, which is not factually consistent with the article. That is, the summary contains information that is not supported by the article.
    
Your task is to generate instructions for editing the summary to make it factually consistent with the article. The correction should preserve most of the summary and only adapt it. Please only make the necessary changes to the summary.

The edit operations are: 1. Remove Information, 2. Add Information, 3. Replace Information, 4. Modify Information, 5. Rewrite Summary 6. Others. Please note that ``Remove Information'' and ``Replace Information'' should be the majority of the operations. ``Add Information'' comes next.

The summary should contain only one sentence. Please keep the style of the summary unchanged, and the length of the summary should be similar before and after the editing.

Example:

Summary: A coalition of US civil rights groups has called on Facebook to do more to protect black people from racist abuse on the social network, saying the site is "racially biased".

The summary is not factually consistent because the source text does not contain the information called on Facebook to do more to protect black people, which is stated in the summary.

Instructions: Replace the information about the claim of the coalition with information about better moderation on the platform.

More instruction examples (summaries omitted):

- ``Instruction 1''

- ``Instruction 2''

- ...

Input:

Article: \{\{Article\}\}

Summary: \{\{Summary\}\}

Please generate the editing instructions for the summary to make it factually consistent with the article:
\end{quote}

As discussed in \S\ref{subsec:llm-critic}, LLMs still lack the ability of \textit{controllable} instruction generation. 
For example, GPT-3.5 responded with ``No editing instructions needed'' 23.9\% of the time, despite being directly instructed to edit a factually inconsistent summary. Conversely, GPT-4 only made such mistakes in 1.8\% of examples.

\end{document}